\documentclass[11pt,a4paper]{article}
\usepackage[hyperref]{AIS2020}
\usepackage{times}
\usepackage{latexsym}

\usepackage{graphicx}
\graphicspath{ {./} }

\usepackage{url}

\title{Integrating Boundary Assembling into a DNN Framework for Named Entity Recognition in Chinese Social Media Text}

\author{Zhaoheng Gong \\
  Harvard Business School \\
  \texttt{zgong@hbs.edu} \\\And
  Ping Chen \\
  University of Massachusetts\\
  Boston\\
  \texttt{Ping.Chen@umb.edu} \\\And
  Jiang Zhou\\
  AI Strike\\
  \texttt{jay.zhou@aistrike.us} \\
}
\date{}

\begin{document}
\maketitle
\begin{abstract}
  Named entity recognition is a challenging task in Natural Language Processing, especially for informal and noisy social media text. Chinese word boundaries are also entity boundaries, therefore, named entity recognition for Chinese text can benefit from word boundary detection, outputted by Chinese word segmentation. Yet Chinese word segmentation poses its own difficulty because it is influenced by several factors, e.g., segmentation criteria, employed algorithm, etc. Dealt improperly, it may generate a cascading failure to the quality of named entity recognition followed. In this paper we integrate a boundary assembling method with the state-of-the-art deep neural network model, and incorporate the updated word boundary information into a conditional random field model for named entity recognition. Our method shows a 2\% absolute improvement over previous state-of-the-art results.
  \end{abstract}

\section{Introduction}

Named entity recognition (NER) is a challenging problem in Natural Language Processing, and often serves as an important step for many popular applications, such as information extraction and question answering. NER requires phrases referring to entities in text be identified and assigned to particular entity types, thus can be naturally modeled as a sequence labeling task. In recent years, a lot of progress has been made on NER by applying sequential models such as conditional random field (CRF) or neural network models such as long short-term memory (LSTM) (e.g., \citealp{mccallum2003early, nadeau2007survey, chiu2016named, lample2016neural}). Yet this task still remains a challenging one, especially in social media domain such as tweets, partially because of informality and noise of such text and low frequencies of distinctive named entities \cite{ritter2011named}.

Chinese is a language that consists of sequential Chinese characters without capitalization information and delimitation between words. Rather than words as in English or Romance languages, the equivalence of an English word in Chinese may contain one or several Chinese characters. Thus Chinese word segmentation is needed as a first step before entity mentions can be recognized. The outputs of Chinese word segmentation are often used as features to support named entity recognition. In the neural network based models, the boundary information can be extracted from hidden layers of a Chinese word segmentation model (e.g., \citealp{peng2016improving, cao2018adversarial}).
\begin{figure*}[ht]
\includegraphics[width=12cm, height=12cm]{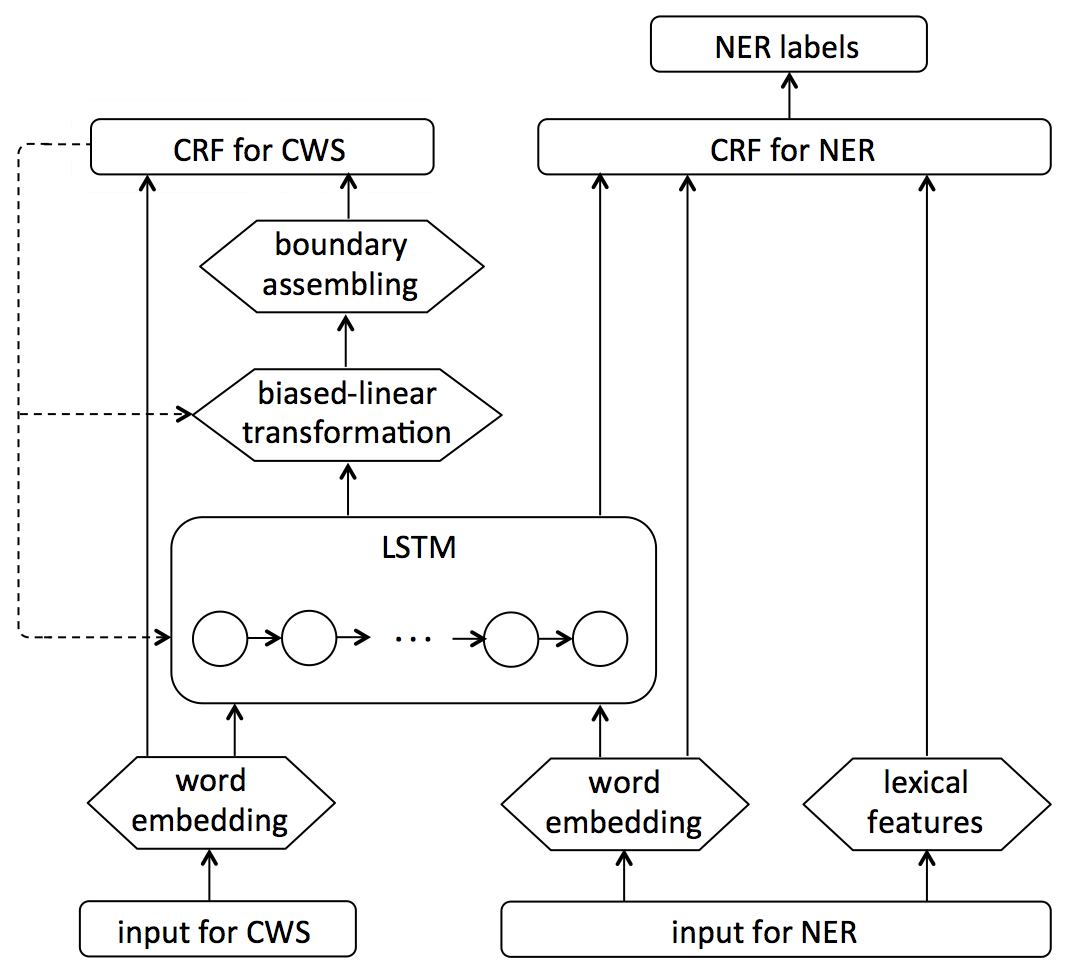}
\centering
\caption{\label{model_diagram} The model for named entity recognition. The LSTM module is trained twice, with inputs first for CWS then for NER. Boundary assembling method is added in between LSTM and CRF for CWS, so that a better representation of segmentation can be obtained. Dashed-line arrows indicate parameter adjustment based on CRF's loss function between each training epoch. CRF for NER takes directly the hidden vectors in LSTM as dynamic features. Abbreviations: CRF: conditional random field; CWS: Chinese word segmentation; NER: named entity recognition; LSTM: long short-term memory.}
\end{figure*}
Relying on outputs of Chinese word segmentation has its own challenge, because Chinese word segmentation is often influenced by the following four factors. First, models trained from corpora in other languages, even if not language-specific such as in \citet{lample2016neural}, cannot be directly applied to Chinese text. Some efforts have been made on Chinese word segmentation recently \cite{chen2015gated, chen2015long, cai2016neural, xu2016dependency}. Second, differences in segmentation criteria exist among labeled corpora. For instance, in the PKU’s People’s Daily corpus, a person's family name and given name is segmented into two tokens, while in the Penn Chinese Treebank corpus they are segmented into one \cite{chen2017adversarial}. Third, for entity mentions that are compound words, models may separate them into fragmented words \cite{chen2015boundary}. Many Chinese words are also morphemes, and separated entity mention fragments may be falsely combined with adjacent words. Fourth, current sequential models perform poorly in identifying named entities with long dependency. Yet many named entities expand over a long range, especially organization names and location names.

Errors caused by Chinese word segmentation can generate a cascading failure, which hurt the performance of downstream tasks such as NER. Latest developments in Chinese NER (e.g., \citealp{peng2015ner, HeS16, peng2016improving, he2017unified, Zhang2018ChineseNU, Zhao2018ChineseNE}) have yet shown little focus on this issue. \citet{chen2015boundary} found that by assembling these single words back together, information groups and sentence structure are better reserved, which could benefit downstream tasks such as NER.

Inspired by \citet{chen2015boundary}, we integrate in this paper a boundary assembling step into the state-of-the-art LSTM model for Chinese word segmentation, and feed the output into a CRF model for NER, resulting in a 2\% absolute improvement on the overall F1 score over current state-of-the-art methods. 

This paper is organized as follows. In Section \ref{model} we discuss our model, which consists of an LSTM module for Chinese word segmentation, a boundary assembling step for more accurate word segmentation, and a CRF module for NER. We show the experiment results and discuss the model performance in Section \ref{exp}. We conclude in Section \ref{conclusion}.

\section{Model}
\label{model}

Our model consists of three modules. A diagram of the model is shown in Figure~\ref{model_diagram}. Characters in the input text for Chinese word segmentation are converted to vectors that are used to train the LSTM module. Output of the LSTM module are transformed by a biased-linear transformation to get likelihood scores of segmentation labeling, then passed through the boundary assembling module. The updated boundary information is used as feature input into the CRF for Chinese word segmentation (CWS), together with character-vector sequences. In each training epoch, CRF for CWS provides feedback into the LSTM hidden layer and the biased-linear transformation to update the hyper-parameters. Another corpus for NER is then used to train the LSTM again, the hidden vector of which (now contains segmentation information updated by the boundary assembling method) is taken as feature input to CRF for NER. Lexical features extracted from the input text for NER, as well as the word embedding sequence, are also taken by the CRF module as input to generate NER labels. This section provides descriptions for each module.

\subsection{LSTM for Word Segmentation}

We choose an LSTM module for the CWS task. Raw input Chinese text is converted from characters to vectors with character-positional input embeddings pre-trained by \citet{peng2016improving} over 112,971,734 Weibo messages using word2vec \citep{mikolov2013distributed}. Detailed parameter settings can be found in \citet{peng2015ner}. The embeddings contain 52,057 unique characters in a 100-dimension space.

The LSTM module takes these vectors into a single layer that contains 150 nodes, and modifies them into likelihood scores for each segmentation label. A biased-linear transformation is carried out on these likelihood scores, generating predicted labels for segmentation. These labels are then modified by the Boundary Assembling module, which we will discuss in detail in the next section. Because labels are in sequence, and dependency may exist among adjacent labels, a transition probability matrix is introduced. The likelihood score together with the transition score are taken by a CRF module with a maximum-likelihood training objective. Feedbacks based on the loss function of the CRF are then given to the LSTM's hidden layer and the biased-linear transformation's parameters for update.

\subsection{Boundary Assembling Method}

In each sentence, Chinese characters are labeled as either Begin, Inside, End, or Singleton (BIES labeling). The likelihood of individual Chinese characters being labeled as each type is calculated by the LSTM module described in the previous section. \citet{chen2015boundary} found in a Chinese corpus that the word label "End" has a better performance than "Begin". This motivates us to carry out a backward greedy search over each sentence's label sequence to identify word boundaries. If two words segmented in a sentence are identified as nouns, and one word is immediately before the other, we assemble their boundaries, creating a new word candidate for entity recognition. This strategy has the advantage to find named entities with long word length. It also reduces the influence caused by different segmentation criteria.

\subsection{CRF for Named Entity Recognition}

A log-bilinear CRF module is used for the NER task, and takes three inputs. The first is the sequential character-positional embeddings mentioned above. The second is the hidden vector from LSTM as dynamic feature inputs. The third is lexical features extracted from the input text. These lexical features are the likelihood of a character being at a specific position of a noun (first character of a noun, second character of a noun, etc.), and is achieved by comparing the character with a pre-defined dictionary trained by \citet{peng2015ner}.

\begin{table*}[ht]
\centering
  \begin{tabular}{|l|l|l|l|l|l|l||l|}
    \hline
    \multicolumn{1}{|c|}{Models} &
    \multicolumn{3}{c|}{Named Entity} &
    \multicolumn{3}{c||}{Nominal Mention} &
    \multicolumn{1}{c|}{Overall} \\
    & Prec & Recall & F1 & Prec & Recall & F1 & F1 \\
    \hline
    \citet{he2017unified} & 61.68 & 48.82 & 54.50 & 74.13 & 53.54 & 62.17 & 58.23 \\
    \hline
    \citet{peng2017Sup4improving} & 66.67 & 47.22 & 55.28 & 74.48 & 54.55 & 62.97 & 58.99 \\
    \hline
    \citet{Xu2018CrossDomainAS} & 59.48 & 54.97 & 57.14 & 72.41 & 53.03 & 61.22 & 59.11 \\
    \hline
    Our method & \textbf{71.33} & 47.22 & 56.82 & 73.89 & \textbf{58.59} & \textbf{65.35} & \textbf{61.06} \\
    \hline
  \end{tabular}
\caption{\label{model_results-table}  The results of two previous models, and results of this study, in which we apply a boundary assembling method. Precision, recall, and F1 scores are shown for both named entity and nominal mention. For both tasks and their overall performance, we outperform the other two models. 
  }
\end{table*}

\section{Experiments}
\label{exp}

\subsection{Datasets}

Datasets used in this study for training, validation, and test are the same as used in Peng et al. \shortcite{peng2016improving} for both word segmentation and named entity recognition. Specifically, dataset for word segmentation is taken from the SIGHAN 2005 bakeoff PKU corpus \cite{emerson2005second}, which includes 123,530 sentences for training and 11,697 sentences for testing. Dataset for named entity recognition is a corpus composed of 1,890 Sina Weibo (a Chinese social media website) messages, with 1,350 messages split into training set, 270 into validation set, and 270 into test set \cite{peng2016improving}. Both named entity and nominal mention are annotated, each with four entity types: person, organization, location, and geo-political entity. A major cleanup and revision of annotations of this corpus has been performed by He and Sun \shortcite{HeS16}. In this study, all results for comparisons are based on this updated corpus.

\subsection{Training Settings}

The weights and hyper-parameters in the LSTM module are adjusted in each iteration using stochastic gradient descent for two stages in tandem. The first stage is based on the CWS input text, and the second stage on NER input text. Since the corpus for CWS is much larger than the corpus for NER, the former corpus is randomly sub-sampled during each training epoch, with each sample containing 13,500 sentences for training step one, and 1,350 sentences for training step two. Models are trained until the F1 score of NER model converges on the validation dataset, or up to 30 epochs.

Models are trained on a Windows PC with a 2.7 GHz Intel Core i7 CPU. For the best performed model, the average training time for the LSTM module is 1897.6 seconds per iteration. Time for data loading, pre-processing, or model evaluation is not included.

\subsection{Results and Discussion}

Our best model performance with its Precision, Recall, and F1 scores on named entity and nominal mention are shown in Table~\ref{model_results-table}. This best model performance is achieved with a dropout rate of 0.1, and a learning rate of 0.05. Our results are compared with state-of-the-art models \cite{he2017unified, peng2017Sup4improving, Xu2018CrossDomainAS} on the same Sina Weibo training and test datasets. Our model shows an absolute improvement of 2\% for the overall F1 score.

This significant improvement validates our method of applying boundary assembling to the segmented sentences, which results in more accurate word segmentation and better semantic understanding. Sentences in the PKU corpus are often segmented into the smallest word units. This results in too fragmented  information and incorrect lexical units when input into a named entity recognition model, although this may benefit some other natural language process tasks. By applying the boundary assembling method, sense group in a sentence is better preserved. The downstream NER model can then take advantage of this, improving its result.

The PKU corpus used for CWS module consists of mainly news articles, which are quite different from the social media Weibo corpus. Performance of an NLP task often drops when tested on a different domain or a corpus of different characteristics. Our improvement indicates that the boundary assembling method is not sensitive to the specific domain, and is a robust method for cross-domain scenarios.

The identification of two nouns that are next to each other depends on the pre-trained lexical features. If our model is tested over out-of-vocabulary dataset, it may not perform well due to the lack of this lexical information.

\section{Conclusion}
\label{conclusion}

In this paper we integrate a boundary assembling step with an LSTM module and a CRF module for Named Entity Recognition in Chinese social media text. With the abundance of social media information, our work is timely and desirable. The improvement in experiment results over existing methods clearly shows the effectiveness of our approach. 

\section*{Acknowledgments}

This research was partially funded by the Engineering Directorate of the National Science Foundation (1820118).

\newpage
\bibliography{AIS2020}
\bibliographystyle{AIS_natbib}

\end{document}